\documentclass[journal,twoside,web]{ieeecolor}
\usepackage{tmi}

\usepackage{amsmath,amssymb,amsfonts}
\usepackage{algorithm}
\usepackage{algpseudocode}

\usepackage{array}
\usepackage{utfsym}
\usepackage{booktabs}
\usepackage[caption=false,font=normalsize,labelfont=sf,textfont=sf]{subfig}
\usepackage{pifont}
\usepackage{makecell}
\usepackage{dsfont}
\usepackage{diagbox}
\usepackage{titlesec}
\usepackage{stfloats}
\usepackage{multirow}
\usepackage{verbatim}
\usepackage{graphicx}
\usepackage{textcomp}
\usepackage{orcidlink}
\usepackage{cite}
\usepackage{url}
\usepackage{color}
\makeatletter
\let\NAT@parse\undefined
\makeatother
\usepackage{hyperref}
\hypersetup{
        hypertex=true,
	colorlinks=true,
	linkcolor=blue,
	filecolor=blue,      
	urlcolor=black,
	citecolor=blue,
}

\def\BibTeX{{\rm B\kern-.05em{\sc i\kern-.025em b}\kern-.08em
    T\kern-.1667em\lower.7ex\hbox{E}\kern-.125emX}}
\begin{document}
\title{Positive Semi-definite Latent Factor Grouping-Boosted Cluster-reasoning Instance Disentangled Learning for WSI Representation}

\author{Chentao Li~\orcidlink{0009-0001-5474-9311}, \IEEEmembership{Graduate Student Member, IEEE}, Behzad Bozorgtabar~\orcidlink{0000-0002-5759-4896}, Yifang Ping~\orcidlink{0000-0002-0699-8401}, Pan Huang~\orcidlink{0000-0001-8158-2628}, \IEEEmembership{Member, IEEE} and Jing Qin~\orcidlink{0000-0002-7059-0929}, \IEEEmembership{Senior Member, IEEE}
\thanks{\textit{Corresponding author: Pan Huang.}}
\thanks{Chentao Li is with the Fu Foundation School of Engineering and Applied Science, Columbia University, New York, NY 10027, USA (e-mail: cl4691@columbia.edu).}
\thanks{Behzad Bozorgtabar is with the Signal Processing Laboratory (LT55),
École Polytechnique Fédérale de Lausanne (EPFL), 1015 Lausanne,
Switzerland, and also with the Radiology Department, Centre Hospitalier Universitaire Vaudois, 1005 Lausanne, Switzerland (e-mail:
behzad.bozorgtabar@epfl.ch).}
\thanks{Yifang Ping is with the Jingfeng Laboratory, Chongqing, China (e-mail: pingyifang@126.com).}
\thanks{Pan Huang is with the Centre for Smart Health, School of Nursing, The Hong Kong Polytechnic University, Hong Kong 999077, SAR China, and also with the Jingfeng Laboratory, Chongqing, China (e-mail: panhuang@polyu.edu.hk).}
\thanks{Jing Qin is with the Centre for Smart Health, School of Nursing, The Hong Kong Polytechnic University, Hong Kong 999077, SAR China (e-mail:harry.qin@polyu.edu.hk).}}
\maketitle

\begin{abstract}
Multiple instance learning (MIL) has been widely used for representing whole-slide pathology images. However, spatial, semantic, and decision entanglements among instances limit its representation and interpretability. To address these challenges, we propose a latent factor grouping-boosted cluster-reasoning instance disentangled learning framework for whole-slide image (WSI) interpretable representation in three phases. First, we introduce a novel positive semi-definite latent factor grouping that maps instances into a latent subspace, effectively mitigating spatial entanglement in MIL. To alleviate semantic entanglement, we employs instance probability counterfactual inference and optimization via cluster-reasoning instance disentangling. Finally, we employ a generalized linear weighted decision via instance effect re-weighting to address decision entanglement. Extensive experiments on multicentre datasets demonstrate that our model outperforms all state-of-the-art models. Moreover, it attains pathologist-aligned interpretability through disentangled representations and a transparent decision-making process. Code and models will be available at \href{https://github.com/Prince-Lee-PathAI/PG-CIDL}{\textit{\textbf{https://github.com/Prince-Lee-PathAI/PG-CIDL}}}.
\end{abstract}


\section{Introduction}\label{intro}
\begin{figure}[!t]
\centering
{\includegraphics[width=3.3in]{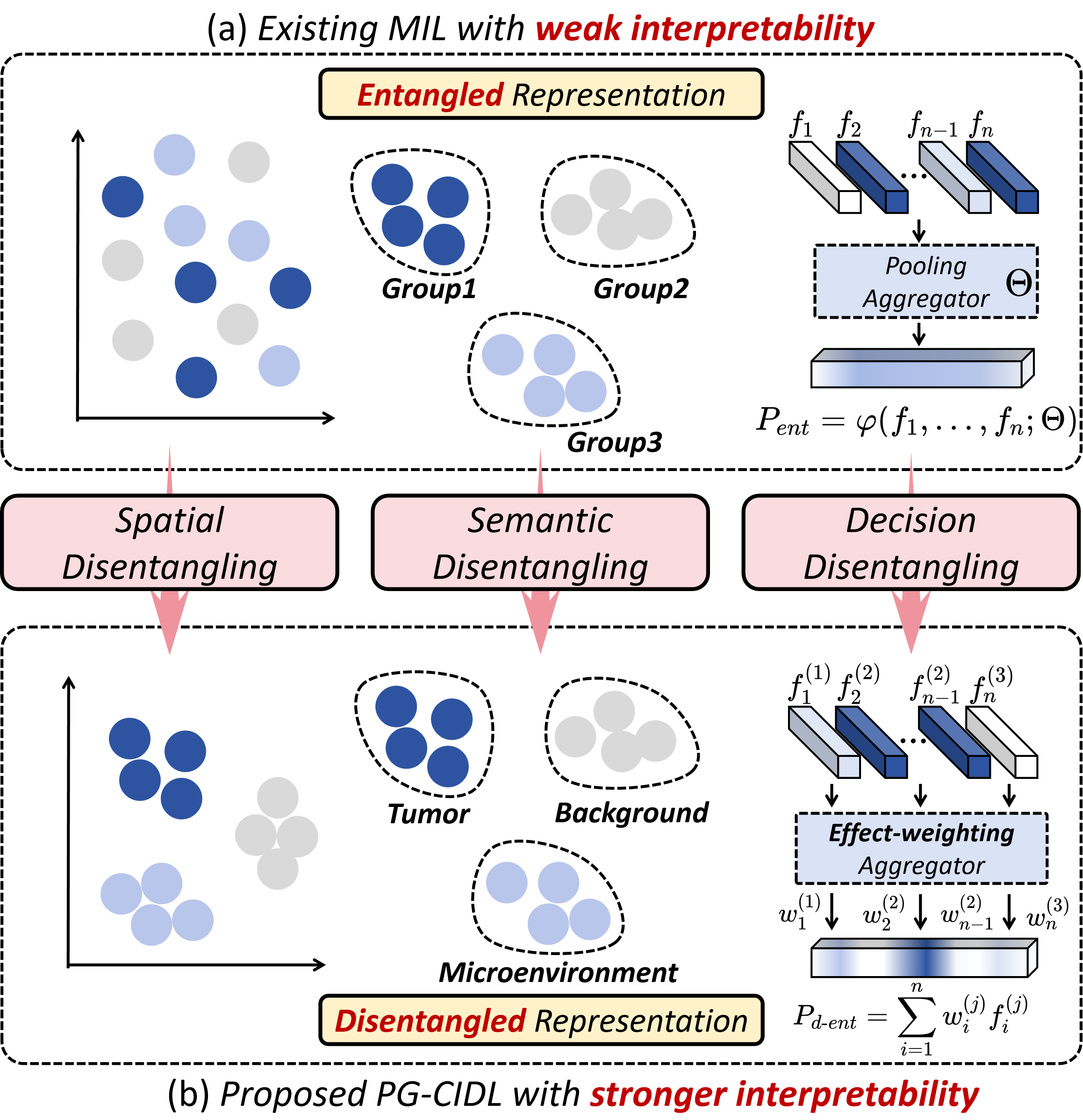}}
\hfil
\caption{Motivation of PG-CIDL. \textbf{(a):} Conventional MIL framework with entangled representation has weak interpretbility. \textbf{(b):} Proposed PG-CIDL framework with spatial, semantic and decision disentanglement has better interpretability.}
\label{fig:fig1}
\end{figure}

\IEEEPARstart{W}{hole} slide images (WSIs), the gold standard for pathological classification, play a vital role in clinical tasks such as diagnosis, prognosis, and metastasis prediction \cite{kumar2020whole}. Multiple instance learning (MIL) has emerged as a promising approach for computational gigapixel WSIs analysis. It is a typical weakly supervised learning framework, where slide-level (bag) labels are available, while abundant instance-level (patch) labels within each slide are few annotated. Existing MIL frameworks suffer from weak interpretability because of entangled instance features from three aspects, \textit{i.e.} spatial entanglement, semantic entanglement and decision entanglement; see Figure \ref{fig:fig1}.

Spatial entanglement arises when spatially adjacent instances exhibit correlated yet distinct pathological factors. It leads the model to confuse their individual contributions due to the absence of explicit spatial constraints. For example, the mixed tumor boundary and inflammatory areas make it difficult to distinguish the pathological border in spatial context. Typically, ABMIL \cite{ilse2018abmil} selects top-k patches as positive instances based on attention scores. ACMIL \cite{zhang2024acmil} extends this by introducing multi-branch attention and stochastic top-k masking to capture diverse informative instances. Including TransMIL \cite{shao2021transmil}, these attention-based MIL framework \cite{li2021dual,ilse2018attention, chen2021multimodal, zhang2024acmil, ilse2018abmil, shao2021transmil} generate attention scores that merely reflect the correlation between instances and bag-level predictions. They are incapable of modeling how instances contribute to pathological spatial patterns \cite{pawlowski2020deepsm}.

Despite spatial disentanglement, the inability to determine the factor group for each instance still hinders the model’s interpretability and representation. From the standpoint of genetic biology, tumor regions are more closely linked to pathological manifestations than non-tumor areas \cite{renshaw2014diagnostic}. However, conventional MIL models under weak supervision lack explicit semantic guidance, thus fusing tumor and non-tumor into unified latent representation. This semantic entanglement fails the model to distinguish which regions are truly responsible for the diagnostic outcome; see Figure \ref{fig:fig1}. Consequently, the learned instance weights fail to relate with pathological factors, which undermines model interpretability and alignment with pathological priors. To this end, cluster-incorporated MILs have been explored in WSI analysis. Panther \cite{song2024fasial} is a prototype-based approach with Gaussian mixture model that summarizes WSI patches into a smaller set of morphological prototypes, transporting into refined WSI features. FuzzyMIL \cite{liu2025fuzzymil} decouples morphological patterns in WSIs through a learnable deep fuzzy clustering framework while cDPMIL \cite{chen2024cdp} models the instance-to-bag structure using a cascade of Dirichlet processes based on feature covariance. These cluster-based models \cite{lu2021clam, jin2023deep, yan2023gcfagg, zheng2024partial, song2024fasial, liu2025fuzzymil, chen2024cdp} merely approximate instance soft labels and suffer from entangled feature representation, which may highlight background noise and degrade both interpretability and classification performance.

With disentangled semantics, the ultimate decision-making process can still be ambiguous. This decision entanglement results from the fuzzy inter-dependent aggregator in MIL, where instance contributions are determined through pooling or attention mechanisms; see Figure \ref{fig:fig1}. This causes the importance of each patch to be influenced by the global feature distribution, blurring the correspondence between attention weights and true pathological relevance. DGR-MIL \cite{zhu2024dgrmil} utilizes a set of global vectors to represent WSI features, while RRT-MIL \cite{tang2024rrtmil} re-embeds degraded features via a regional transformer. Causality-based MIL models have been explored to capture instance correlations during decision-making \cite{wu2024causalACML,guo2024causalityTIP, nan2024establishing, mfcmil2025,lin2024cimil, lin2023ibmil, chen2024camil,mfcmil2025}. For example, IBMIL \cite{lin2023ibmil} addresses the spurious correlations between bags and labels based on the backdoor adjustment, while CaMIL \cite{chen2024camil} blocks the spurious association between disease and the color by applying the front-door adjustment. MFC-MIL \cite{mfcmil2025} employs an adaptive memory module to simulates causal interventions to mitigate confounders in diagnosis tasks. In general, although these methods tend to refine WSI representation, not only the final decision-making remains ambiguous without specific causal effect weighting, but they lack structural causal model to build disentangled relationships.

To address these entanglement, we introduce PG-CIDL, a three-phase disentangled representation learning framework for interpretable WSI analysis, which enables the model to disentangle spatially, semantically and decision ambiguous representations; see Figure \ref{fig:fig1}. 
First, we proposed a novel positive semi-definite latent factor grouping (PSD-LFG) method, which maps spatially entangled instances into a implicit subspace and adaptively partition them into three latent groups. To address semantic entanglement, we develop the cluster-reasoning instance disentangling (CID) in the second phase. By instance probability counterfactual inference, we measure its causal effect, for which features are disentangled by the extent of effects. Finally, the obtained effects are used to re-weighting original feature to get a refined WSI representation via generalized linear weighted summation, mitigating decision entanglement. Through end to end training, our proposed PG-CIDL not only disentangles instances into domain-specific semantics, but also improves the classification performance and interpretability. Our contributions can be summarized as follows:

\begin{itemize}
    \item We proposed positive semi-definite latent factor grouping for spatial disentanglement, which maps features into latent subspace and provide robust grouping results.
    \item We introduce cluster-reasoning instance disentangling to identify groups into separate semantics with effects measured by instance probability counterfactual inference.
    \item To address decision entanglement, we utilize instance effect re-weighting in a generalized linear weighted decision to refine the WSI representation.

\end{itemize}

\section{Methodology}

\subsection{Problem Statement}
\noindent \textbf{Assumption:} \textbf{(1)}: Tumor, microenvironment, and irrelevant background can coexist within each WSI. These weakly-supervised patches meet the criteria for adopting the MIL framework. \textbf{(2)}: Factors hold causal dependencies in disentangled representation learning (DRL), rather than independence. \textbf{(3)}: Tumor cells are directly related with pathological grading, while microenvironment interacts with tumor cells, jointly shaping the grading outcome.

\noindent \textbf{MIL Formulation:} The goal of WSI-based classification task in MIL framework is to learn a permutation-invariant scoring function \cite{zaheer2017deep} $\mathcal{F}$ that maps the set of instances $\mathbf{X}=\{\boldsymbol{x}_1,\dots,\boldsymbol{x}_n\}$ to label $\mathbf{Y}$. The prediction process can be formulated as follows:
\begin{equation}
    \hat{\mathbf{Y}}=\mathcal{F}(\mathbf{X}) \quad\text{where}\quad \mathcal{F}(\mathbf{X}) \equiv f\bigl(\sigma(\mathcal{A}(\mathbf{X})\bigr)
\end{equation}
where $\sigma(\cdot),f(\cdot)$ denote the aggregation function, prediction head with softmax normalization, respectively. $\mathcal{A}(\cdot)$ denotes the feature extractor.

WSI-based MIL framework suffer from weak interpretability because of spatial, semantic and decision entanglement. Therefore, we apply DRL to address these entangled dependencies \cite{wang2024DRL}. 

\noindent \textbf{DRL Formulation:} Given the set of instances $\mathbf{X}=\{\boldsymbol{x}_1,\dots,\boldsymbol{x}_n\}$, we assume factor tumor (\(\mathbf{T}\)), microenvironment (\(\mathbf{E}\)) and background noise (\(\epsilon\)) ambiguously generates the whole distribution by function $\boldsymbol{\Phi}(\cdot)$. Thus, DRL aims to disentangle this mapping from given instances to hidden factors $\mathbf{F}=\{\mathbf{h}_1:=\mathbf{T},\mathbf{h}_2:=\mathbf{E},\mathbf{h}_3:=\epsilon\}$, \textit{i.e.},

\begin{equation}
    \mathbf{X}=\boldsymbol{\Phi}(\mathbf{F}) \quad \rightarrow \quad \mathbf{h}_j=\boldsymbol{\Phi}^{-1}(\boldsymbol{x}_i) 
\end{equation}
In this case, spatial disentanglement can be modeled by expanding the distance $\mathcal{D}$ between each two factors, \textit{i.e.}, maximizing the following,

\begin{equation}
\begin{aligned}
     \max \limits_{\mathcal{A},\boldsymbol{\Phi}} \mathcal{D}=\sum_{j\neq j'} \Big\|\mathbb{E}_{\boldsymbol{x\sim \mathcal{X}}} \Big[ \mathds{1}_{\mathbf{h}_j} \Big(\boldsymbol{\Phi}^{-1}(\boldsymbol{x})\Big)\mathcal{A}(\boldsymbol{x}) \Big]\\
     - \mathbb{E}_{\boldsymbol{x'\sim \mathcal{X}}} \Big[ \mathds{1}_{\mathbf{h}_{j'}} \Big(\boldsymbol{\Phi}^{-1}(\boldsymbol{x'})\Big)\mathcal{A}(\boldsymbol{x'}) \Big] \Big \|_2
\end{aligned}
\end{equation}

By the assumption, \textit{factors of variation} in pathological grading are not independent semantics but hold certain causal relations. To address semantic entanglement, we introduce two decoupled SCMs to characterizes these disentangling factors in DRL as prior knowledge. In Figure \ref{fig:scm}, both two SCMs meet the back-door criterion, which provides a theory for using Pearl's back-door adjustment to estimate the causal effect of one factor by controlling another; see \eqref{eq:pearl}. Thus, it guarantees the feasibility of PG-CIDL model desgin.

\begin{figure*}[!t]
\centering
{\includegraphics[width=7.2in]{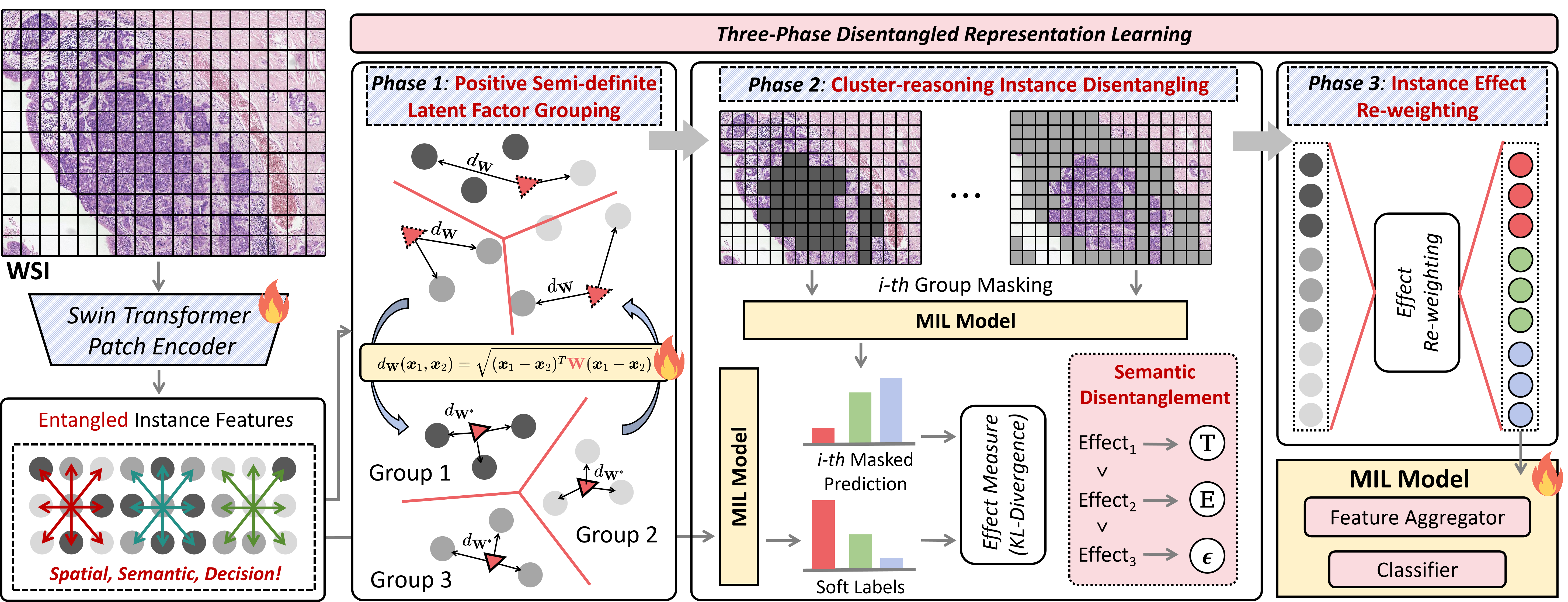}}
\hfil
\caption{Overview of \textbf{PG-CIDL}. Entangled Features extracted by patch encoder are categorized into three groups via \underline{\textbf{P}}ositive semi-definite latent factor \underline{\textbf{G}}rouping. They are identified into tumor, microenvironment and background factors by \underline{\textbf{C}}luster-reasoning \underline{\textbf{I}}nstance \underline{\textbf{D}}isentangling. End-to-end optimization is performed across all phases for disentangled representation \underline{\textbf{L}}earning with instance effect re-weighted features.}
\label{fig:fig2}
\end{figure*}

\begin{figure}[!t]
\centering
\includegraphics[width=3.5in]{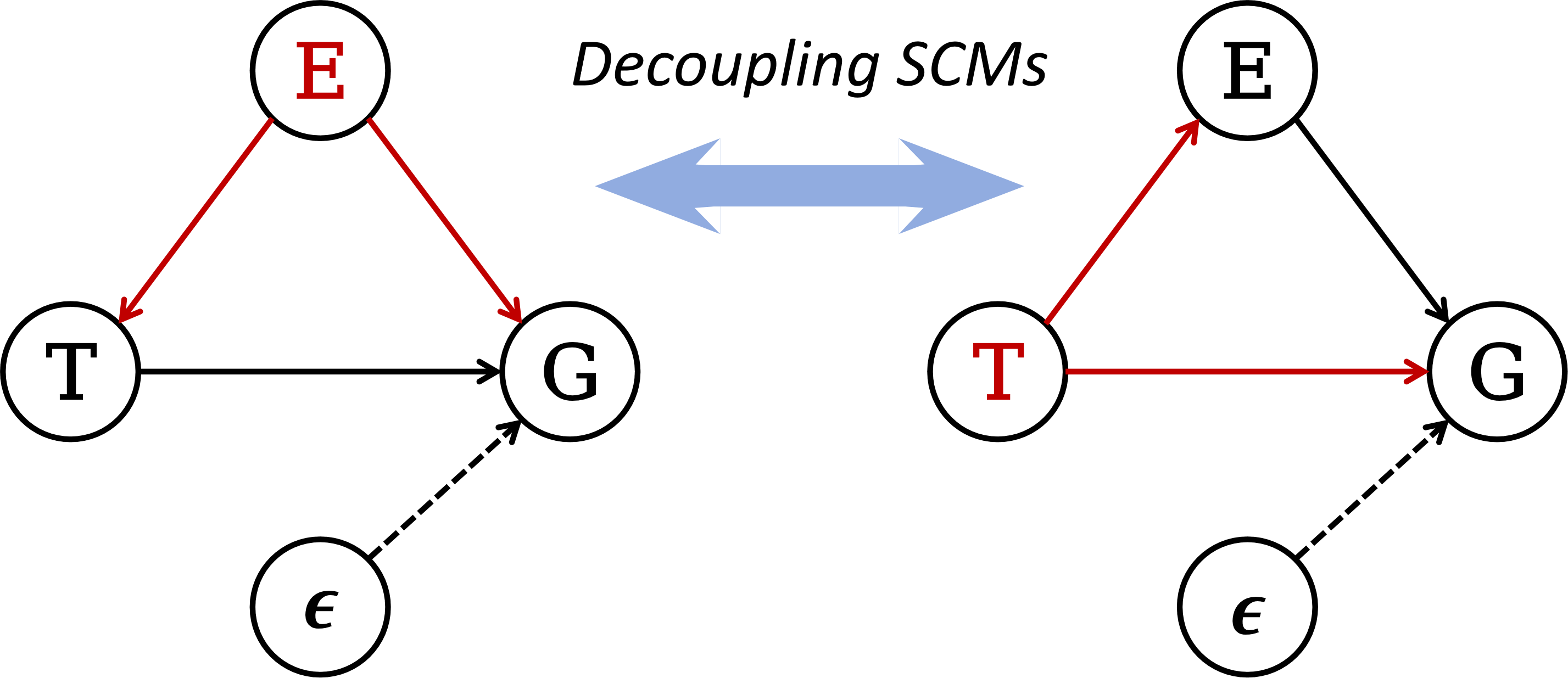}
\caption{Two decoupled SCMs for disentangling factors in DRL, where node T is for factor tumor, node E for microenvironment, node G for pathological grading outcome and node \(\epsilon\) for possible background noise. \label{fig:scm}}
\end{figure}

\begin{equation}\label{eq:pearl}
    P\bigl(\mathbf{G}\mid\mathsf{do}(\mathbf{T})\bigr)
=\sum_k P\bigl(\mathbf{G}\mid \mathbf{T},\,\mathbf{E}=e_k\bigr)\,P(\mathbf{E}=e_k)
\end{equation}
where $e_k$ denotes the controlling option of $k$ instances belonging to factor $\mathbf{E}$.

Furthermore, we employ information entropy to determine the decision disentanglement.  Let $\{\mathbf{x}_1,\dots,\mathbf{x}_n\}$ be random variables of instance in one bag. Then the following theorem holds \cite{shao2021transmil}:

\textbf{Theorem 1}. \textit{The joint information entropy can be expressed as $\sum_{i=1}^{n}H(\mathbf{x}_i)$ iff. $\mathbf{x}_i$ are i.d.d. variables. Furthermore, when they are not i.d.d., we can derive that:}
\begin{equation}
    \begin{aligned}
    H(\mathbf{x}_1,\cdots,\mathbf{x}_n)&= H(\mathbf{x}_1) + \sum_{i=2}^{n}H(\mathbf{x}_i|\boldsymbol{\Phi}^{-1}(\mathbf{x}_1,\cdots,\mathbf{x}_{i-1})\\
        &\le \sum_{i=1}^{n}H(\mathbf{x}_i)
    \end{aligned}
\end{equation}

Assume that these variables are disentangled into three factors tumor, microenvironment, and background. Let bag $\mathbf{X}$ be partitioned into three subsets with sizes $n_1$, $n_2-n_1$ and $n-n_2$. We can derive that:
\begin{equation}\label{eq3}
    \begin{aligned}
        H^*(\mathbf{x}_1,&\cdots,\mathbf{x}_n)= w_1\sum_{i=1}^{n_1}H(\mathbf{x}_i | \boldsymbol{\Phi}^{-1}(\mathbf{x}_i)\in \mathbf{h}_1=\mathbf{T})\\
        &+w_2\sum_{i=n_1+1}^{n_2}H(\mathbf{x}_i|\boldsymbol{\Phi}^{-1}(\mathbf{x}_i)\in \mathbf{h}_2=\mathbf{E}) \\ 
        &+w_3\sum_{i=n_2+1}^{n}H(\mathbf{x}_i|\boldsymbol{\Phi}^{-1}(\mathbf{x}_i)\in \mathbf{h}_3=\epsilon)\\
        &\le H(\mathbf{x}_1)+\sum_{i=2}^{n}H(\mathbf{x}_i|\boldsymbol{\Phi}^{-1}(\mathbf{x}_1,\cdots,\mathbf{x}_{i-1})) \\
        &\le \sum_{i=1}^{n}H(\mathbf{x}_i)
    \end{aligned}
\end{equation}
\noindent where $H^*$ denotes the correlated information entropy and $w_j$ denotes the instance effects. The weighted summation after semantic disentanglement only decrease the entropy value compared to the original fuzzy decision. Therefore, the total information entropy of a bag in \eqref{eq3} decreases, which enhance the ability to mitigate decision entanglement.

\subsection{Overview of PG-CIDL}
The overall pipeline of PG-CIDL is illustrated in Figure \ref{fig:fig2}. The cropped patches are extracted into feature representations by learnable Swin Transformer \cite{liu2021swin}. To improve the interpretability of WSI representation within spatially, semantically and decision entangled features, we apply a three-phase disentangled representation learning framework. In phase 1, we perform positive semi-definite latent factor grouping to obtain three semantically ambiguous factors of variation. In phase 2, we disentangle these factors based on instance probability counterfactual inference, where the estimated causal effects categorize them into tumor, microenvironment and background. In phase 3, we refine the original representation by instance effect re-weighting.

\subsection{Positive Semi-definite Latent Factor Grouping}
Classic clustering methods with $L_p$ norm with equal weights disregard the interaction and fusion between each two dimensions. As a result, they exacerbate the spatial entanglement of instance features, leading to semantic ambiguity and reduced interpretability. Inspired by the idea of metric learning and deep clustering \cite{caron2018deep, tang2020unsupervised}, we proposed a latent factor grouping approach with positive semi-definite constraint to address the limitations.

The general form of metric learning is based on Mahalanobis distance in the following expression.
\begin{equation}\label{eq6}
    \begin{aligned}
        d_{\mathbf{W}}(\boldsymbol{x}_1,\boldsymbol{x}_2)=\sqrt{(\boldsymbol{x}_1-\boldsymbol{x}_2)^ \top \mathbf{W}(\boldsymbol{x}_1-\boldsymbol{x}_2})
    \end{aligned}
\end{equation}
where $\mathbf{W}\in \mathbb{R}^{n\times n}$ is a trainable positive semi-definite matrix and $\boldsymbol{x}_1, \boldsymbol{x}_2 \in \mathbb{R}^{n\times 1}$ denote the instance feature representation vectors. If $\mathbf{W}$ is an identity matrix, the distance degrades as standard Euclidean distance. 

Positive semi-definite matrix ensures the legality and non-negativity of the metric and facilitates robust metric learning process. Furthermore, compared to diagonal matrix, a general positive semi-definite matrix models more interaction effects among pathological factors, while a diagonal matrix only adjusts the weight of each feature independently.

\textbf{Fact 1}. \textit{For any $\mathbf{A}\in \mathbb{R}^{n\times n}$, the matrix $\mathbf{A}^\top\mathbf{A}$ is always positive semi-definite.}

Therefore, we replace the $L_p$ norm with the adaptive distance $d_{\mathbf{W}}(\cdot,\cdot)$ and ensure \(\mathbf{W}\succeq0\) by parameterizing \(\mathbf{W} = \mathbf{A}^\top \mathbf{A}\), \(\mathbf{A}\in\mathbb{R}^{r\times n}\), where $r<n$ ensures a low-rank structure for easier computation. By incorporating optimal \(\mathbf{W}^*=\mathbf{A}^{*\top}\mathbf{A}^*\) into KMeans ($k=3$, Figure \ref{fig:fig2}) we can also assume that PSD-LFG maps the distance into a latent subspace in \eqref{eqA}, where we can perform more robust and accurate clustering results, thus alleviating spatial entanglement. 

\begin{equation}\label{eqA}
    \begin{aligned}
        d_{\mathbf{W}^*}(\boldsymbol{x}_1,\boldsymbol{x}_2)&=\sqrt{(\boldsymbol{x}_1-\boldsymbol{x}_2)^ \top \mathbf{A{^*}^{\top}A}^*(\boldsymbol{x}_1-\boldsymbol{x}_2})\\
        =&\|\mathbf{A}^*(\boldsymbol{x}_1-\boldsymbol{x}_2)\|_2
    \end{aligned}
\end{equation}

\begin{algorithm}[!t]
\small
\caption{Cluster-reasoning Instance Disentangling and Instance Effect Re-weighting in PG-CIDL}
\label{alg:causal-cluster}
\begin{algorithmic}[1]
\Require Instance groups \(\mathbf{Z}=\{\mathbf{Z}_1,\mathbf{Z}_2,\mathbf{Z}_3\}\); each \(\mathbf{Z}_k\) belongs to one of the factors TC, ME and BG.
\Ensure  Final bag feature \(\mathbf{Z}_{\rm final}\) and factor map \(\texttt{factor}\)
\State \(P_v \gets f\bigl(\sigma(\mathbf{Z})\bigr)\)  \Comment{vanilla prediction}
\For{\(k=1\) to \(3\)}
  \State \(\mathbf{Z}_{\setminus k} \gets \bigl\{\mathbf{Z}_j : j\neq k\bigr\}\)
  \State \(P_k \;\gets\; f\bigl(\sigma(\mathbf{Z}_{\setminus k})\bigr)\)
  \State \(D_k \;\gets\; D_{\rm KL}\bigl(P_v \Vert P_k\bigr)\)
\EndFor
\State \(\{w_k\}_{k=1}^3 \gets \mathtt{Normalize}(\{D_k\})\)
\State \(\texttt{order}\gets\mathtt{argsort}(\{w_k\},\,\texttt{descend})\)
\State \(\texttt{factor}[\texttt{TC}]\gets \mathbf{Z}_{\texttt{order}[1]},\;\dots,\;\)
  \(\texttt{factor}[\texttt{BG}]\gets \mathbf{Z}_{\texttt{factor}[3]}\)
\State \(\mathbf{Z}_{\rm final}\gets \texttt{mean}\bigl(\sum_{k=1}^3 w_k\,\mathbf{Z}_k\bigr)\)
\State \Return \(\mathbf{Z}_{\rm final},\,\texttt{factor}\)
\end{algorithmic}
\end{algorithm}

\subsection{Cluster-reasoning Instance Disentangling}
Based on the results from Phase 1, we obtained three latent yet semantically entangled instance groups. According to the SCMs introduced before as prior knowledge, we proposed CID, which leverages counterfactual inference to estimate the causal effects within factors, thereby disentangling ambiguous semantics and improving the model’s interpretability and generalization.
Figure \ref{fig:fig2} and Algorithm \ref{alg:causal-cluster} intuitively present the idea of CID, which involves the following three steps.

\noindent \textbf{Group Masking:} Let $\mathcal{A}(\mathbf{X})=\mathbf{Z}\rightarrow\{\mathbf{Z}_1,\mathbf{Z}_2,\mathbf{Z}_3\}$ be three instance groups from phase 1. The intervention sets $\mathsf{do}(\mathbf{Z}_{\setminus k})$ as
\begin{equation}
    \mathbf{Z}_{\setminus k}=\mathbf{Z}\setminus \mathbf{Z}_k, \quad k=1,2,3
\end{equation}
This masking severs all back-door paths through $\mathbf{Z}_k$ and thus isolates its causal effect. Then the prediction under that intervention is
\begin{equation}
    P_k=f(\sigma(\mathbf{Z|} \mathsf{do}(\mathbf{Z}_{\setminus k})))
\end{equation}
where we consider the counterfactual of factor $\mathbf{Z}_k$ as $\mathbf{Z}_{\setminus k}$ that masking $\mathbf{Z}_k$. This masked prediction serve as an important component for causal effect measurement.

\begin{table*}[!t]
\small
\centering
\renewcommand\arraystretch{1.1}
\caption{Performance comparison of WSI classification on multicentre datasets with Swin-T pretrained features.\label{tab:table1}}
\begin{tabular}{lcccccccccccc}
\Xhline{0.8pt}
\multirow{2}*{Models} & \multicolumn{3}{c}{AMU-CSCC} & \multicolumn{3}{c}{AMU-LSCC}& \multicolumn{3}{c}{CAMELYON16} & \multicolumn{3}{c}{DHMC-LUNG} \\ \cmidrule(r){2-4} \cmidrule(r){5-7} \cmidrule(r){8-10} \cmidrule(r){11-13}
~& \makebox[0.06\textwidth][c]{ACC} &\makebox[0\textwidth][c]{WF1} & \makebox[0\textwidth][c]{AUC} & \makebox[0\textwidth][c]{ACC} &
\makebox[0\textwidth][c]{WF1} & \makebox[0\textwidth][c]{AUC} & \makebox[0\textwidth][c]{ACC} &
\makebox[0\textwidth][c]{WF1} & \makebox[0\textwidth][c]{AUC} & \makebox[0\textwidth][c]{ACC} &
\makebox[0\textwidth][c]{WF1} & \makebox[0\textwidth][c]{AUC} \\ \hline
 ABMIL & 0.8302 & 0.8073 & 0.9547 &0.6739&0.6701&0.7888&0.7429&0.7314&0.7918 &0.7333&0.7333&\underline{0.8123}\\
CLAM-SB & 0.8302 & 0.7974 & 0.9581&0.6957&0.6933&0.8367&0.7429&0.7229&0.7478 &0.7556&0.7551&0.7783\\
CLAM-MB  & 0.8774 & 0.8788 & 0.9336&0.6812&0.6811&0.8031&0.7429&0.7314&0.7339 &0.7333&0.7333&0.7684\\
TransMIL & 0.8868 & 0.8860 & 0.9521&0.8261&0.8276&0.9021&0.7429&0.7437&0.6898 &0.7333&0.7333&0.8084\\
DTFD-MIL & 0.8113 & 0.7682 & 0.9036 &0.5435&0.5186&0.7245&0.7143&0.6972&0.7306 &0.6889&0.6671&0.6983\\
IBMIL-DS  & 0.7830 & 0.7421 & 0.8881&0.6377&0.6374&0.7672&0.7429&0.7437&0.7037  &0.6889&0.6889&0.7091\\
ILRA-MIL & 0.8491 & 0.8423 & 0.9123&0.6884&0.6792&0.8048&0.7714&0.7687&0.7306 &0.7778&0.7767&0.8074\\
S4MIL & 0.8491 & 0.8163 & 0.9605 &0.7971&0.7970&0.8845&0.7429&0.7429&0.7322 &0.7111&0.7052&0.7007\\
FRMIL & 0.8113 & 0.7668 & 0.8990&0.6377&0.6383&0.8020&0.7143&0.6972&0.6914 &0.7111&0.7114&0.7442\\
DGR-MIL   & 0.8868 & 0.8858 & 0.9613&0.8188&0.8192&0.9147&\underline{0.8286}&\underline{0.8265}&\underline{0.8253} &\underline{0.8000}&\underline{0.8002}&0.7886\\
ACMIL-MHA & \underline{0.9057} & \underline{0.9070} &0.9701&\underline{0.8551}&\underline{0.8568}&\underline{0.9219}  &0.8000&0.8007&0.7331 &0.6889&0.6671&0.6849\\ 
RRT-MIL & 0.8585 & 0.8560 & 0.9265 &0.7681&0.7684&0.8732&0.7143&0.7109&0.7608 &0.7778&0.7746&0.7521\\ 
MFC-MIL & 0.8962 & 0.8940 & \underline{0.9723} &0.8261&0.8268&0.9126&0.7714&0.7714&0.7208 &0.7778&0.7767&0.7931\\ \hline
\textbf{ PG-CIDL (ours)} & \textbf{0.9434} & \textbf{0.9450} & \textbf{0.9859}&\textbf{0.9275}&\textbf{0.9264}&\textbf{0.9719}&\textbf{0.8857}&\textbf{0.8821}&\textbf{0.9282} &\textbf{0.8444}&\textbf{0.8413}&\textbf{0.8820}\\
\Xhline{0.8pt}
\end{tabular}
\end{table*}

\noindent \textbf{Effect Measurement:} Let $P_v=f(\sigma(\mathbf{Z})$ be the vanilla prediction without any intervention, which serves as the reference and soft labels for comparison. We regard $P_v$ as the target distribution and $P_k$ as the ones approximate the target. Due to this asymmetry, we employ Kullback-Leibler divergence to calculate pairwise difference $D_k$ as effects in \eqref{eq7}.
\begin{equation}\label{eq7}
    D_k=D_{KL}(P_v||P_k)=\sum_z P_v(z) \log(\frac{P_v(z)}{P_k(z)})
\end{equation}
where $D_k$ reveals how much the masked group of instances influence the original prediction. 

\noindent \textbf{Semantic Disentangling:} A larger $D_k$ suggests a more pathologically important group (e.g. tumor), while a smaller one indicates irrelevant groups (e.g. background and microenvironment). Consequently, we can causally identify the \emph{factors of variation} of each group by tumor, microenvironment and background based on the previous assumption; see Algorithm \ref{alg:causal-cluster}.

\subsection{Instance Effect Re-weighting and Optimization}
We normalize each “leave‑one‑out” prediction divergence \(D_k\) to get approximate effect \(w_k = \frac{D_k}{\sum_j D_j}\) and perform effects re-weighting to refine the representation.
\begin{equation}
    \mathbf{Z}_{\text{final}}=\texttt{mean}\bigl(\sum_{k=1}^3 w_k\,\mathbf{Z}_k\bigr)
\end{equation}
which can reflect how the previously disentangled instances effect the final WSI representation and mitigating decision entanglement. Finally, we perform end-to-end optimization across all stages to learn more task-related, interpretable and generalized feature representations. Specifically, the loss function is formulated as follows:
\begin{equation}
    \begin{aligned}
        &\mathcal{L}=\mathcal{L}_{ce}(f(\sigma(\mathbf{Z_{\text{final}}})))-\gamma \cdot d_{\text{reg}}\\
        &d_{\text{reg}}=d_{\mathbf{W}}\Big(\text{mean}(\mathbf{Z}^{\text{TC}}),\text{mean}(\mathbf{Z}\setminus\mathbf{Z}^{\text{TC}})\Big)
    \end{aligned}
\end{equation}
where $\mathcal{L}_{ce}$ denotes the cross entropy. Since KMeans algorithm is not normally differentiable, this group-separation regularizer $d_{\text{reg}}$ that encourages the mean of the tumor group to move away from the mean of all other instances helps to optimize the trainable matrix $\mathbf{W}$ by computing the gradients \(\frac{\partial \mathcal{L}}{\partial \mathbf{A}}\).

\begin{figure*}[!t]
\centering
{\includegraphics[width=7.3in]{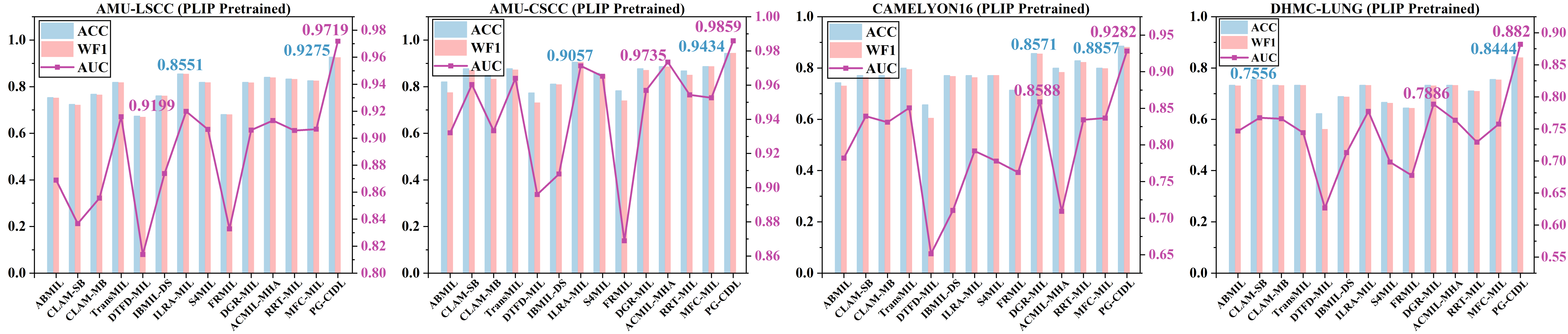}}
\hfil
\caption{Performance comparison on multicentre datasets with PLIP pretrained features.}
\label{fig:fig4}
\end{figure*}

\begin{figure*}[!t]
\centering
{\includegraphics[width=6.5in]{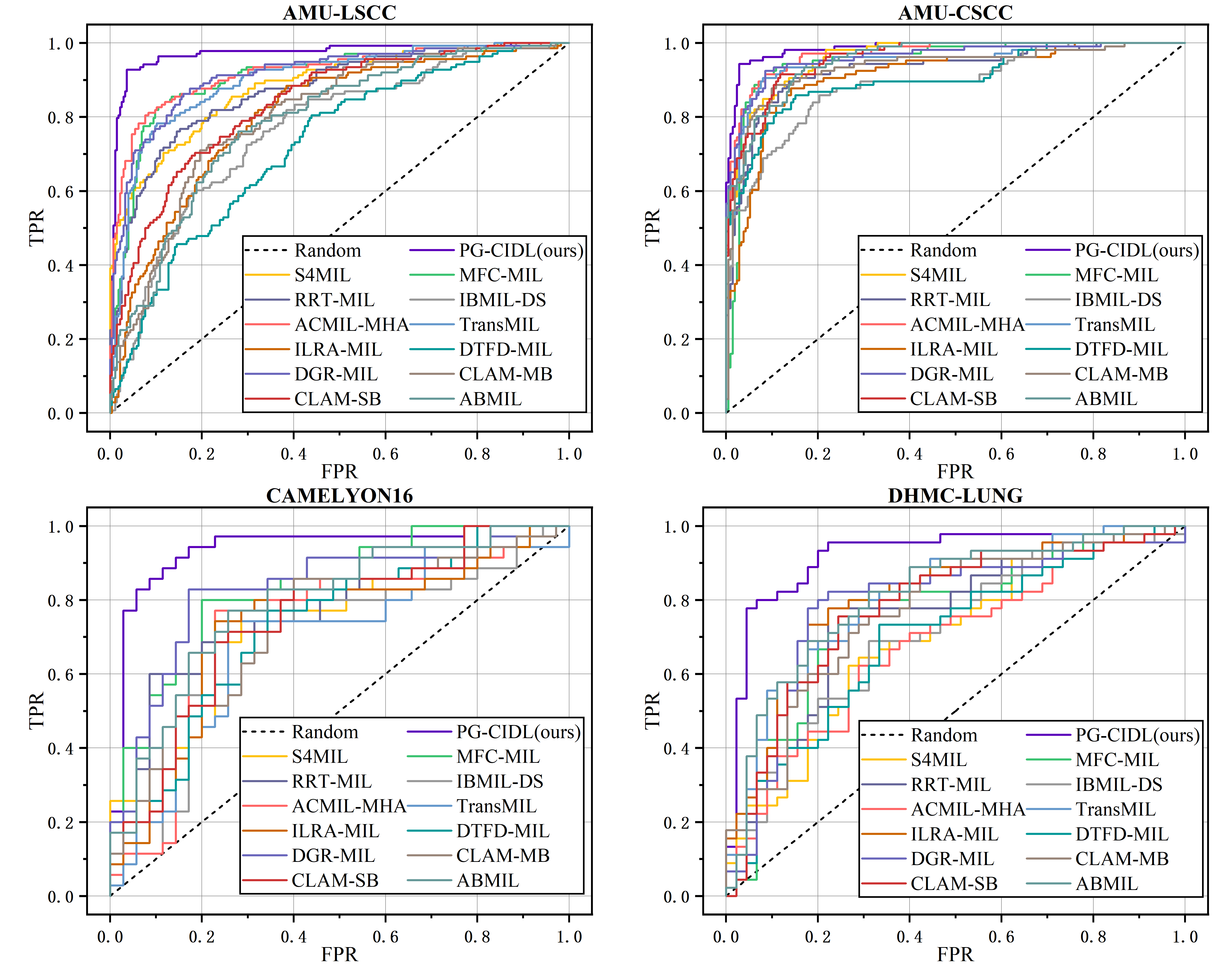}}
\hfil
\caption{ROC plots' comparison on multicentre datasets.}
\label{fig:roc}
\end{figure*}





\section{Experimental Results}
\subsection{Datasets and Setup}
\noindent \textbf{Datasets:} To validate PG-CIDL for diagnosis and sub-typing tasks, we conduct extensive experiments on two public datasets and two private datasets. Two private WSI datasets were collected from Army Medical University, where \textbf{AMU-LSCC} is for laryngeal squamous cell carcinoma pathological grading and \textbf{AMU-CSCC} is for cervical squamous cell carcinom. AMU-LSCC dataset includes 342 whole slide images that was divided by a 6:4 training-validation ratio for each category. Specially Grade I contains a total of 89 WSIs, Grade II contains 152 WSIs and Grade III includes 101 WSIs. AMU-CSCC dataset includes 262 whole slide images, where Grade I contains a total of 27 WSIs, Grade II contains 127 WSIs and Grade III includes 108 WSIs. Two chosen public datasets are \textbf{CAMELYON16} for breast cancer lymph node metastasis detection \cite{bejnordi2017camelyon16} and \textbf{DHMC-LUNG} \cite{wei2019lung} for lung adenocarcinoma classification.

\noindent \textbf{Compared MIL Models:} We implemented eleven state-of-the-art (SOTA) MIL models for comparison. They are ABMIL \cite{ilse2018abmil}, CLAMs \cite{lu2021clam}, TransMIL \cite{shao2021transmil}, DTFD-MIL \cite{zhang2022dtfdmil}, ILRA-MIL \cite{xiang2023ilramil}, IBMIL \cite{lin2023ibmil}, S4MIL \cite{fillioux2023s4mil}, DGRMIL \cite{zhu2024dgrmil}, FRMIL \cite{chikontwe2024frmil}, ACMIL \cite{zhang2024acmil}, RRT-MIL \cite{tang2024rrtmil} and MFC-MIL \cite{mfcmil2025}. Moreover, we report Accuracy, Area Under Curve (AUC), and Weighted F1-score for evaluation metrics.

\noindent \textbf{Implementation:} For comprehensive comparison, we extract features with Swin Transformer \cite{liu2021swin} pretrained on ImageNet-11k and PLIP \cite{huang2023plip} pretrained on OpenPath, respectively. We utilize Rmsprop optimizer for training. Models on two private datasets were trained with a batch size of 2 and 100 epochs. For public datasets, batch size is set 1. The random seed was fixed 0 across all stages. The input resolution of PG-CIDL is $96\times 96 \times3$. The learning rate schedule was: $1\times10^{-5}$ for epochs 1–50, $5 \times 10^{-6}$ for epochs 51–75, and $1 \times 10^{-6}$ for epochs 76–100. The hyper-parameter $\gamma$ in loss function was set 0.1. Patches were grid cropped from each WSI in 10x magnification and we filtered out instances with blank areas. 

\noindent \textbf{Computational Specs:} Experiments were conducted on Ubuntu 22.04 with x86\_64 architecture with four \textit{NVIDIA A10 Tensor Core 24GB}. For computational framework and library versions: we use PyTorch 2.6.0, CUDA 12.4, cuDNN 9.1. For training CAMELYON16 specifically, the peak memory usage is 13GB each GPU at one batch. In terms of model inference, it takes 0.123s per WSI for feature extraction and 0.355s for three-phase DRL.

\subsection{Comparison with SOTA Methods}
For diagnostic tasks like CAMELYON16 (tumor vs. normal), WSIs may exhibit varying differentiation states, but rather than discussed factors. Yet, they can be more accurately diagnosed when patches are categorized with causal relevance. Table \ref{tab:table1} and Figure \ref{fig:fig4} present the performance of SOTA methods on multicentre datasets.

The results suggest that our PG-CIDL outperforms all SOTA models across all three evaluation metrics and four benchmarks. For features pretrained on ImageNet-11k, while ACMIL-MHA achieves the second-best classification performance on AMU-LSCC, our model improves the ACC and AUC by 7.24\% and 5.00\%, respectively. On CAMELYON16 dataset, the improvement is 5.71\% and 10.29\% compared to DGR-MIL. AMU-CSCC and DHMC-LUNG have also seen the similar trend that our proposed PG-CIDL achieves the best classification performance. Although two private datasets suffer from class imbalance, the reported Weighted-F1 scores also show the similar improvement with 3.80\% and 6.96\% of gains respectively. Figure \ref{fig:roc} illustrates the curve of PG-CIDL lies above all the other curves and suggests the largest area under the curve (AUC), which implies that our PG-CIDL achieves the highest predictive confidence among all SOTA models.

In Figure \ref{fig:fig4}, we notice that incorporating features extracted by PLIP pretrained on WSIs,  most models fail to exhibit a significant improvement. Specifically, we only observe a moderate increase on CAMELYON16, with the second-best ACC rising from 82.86\% to 85.71\% and AUC from 82.53\% to 85.88\%. For detailed information, please refer to the Supplementary Material. Despite this, our PG-CIDL remains superior to all compared SOTA models. We attribute this achievement to the advantages of refined representation from PSD-LFG and CID in our disentangled representation learning framework, which will be validated in the following sections.

\begin{table}[!t]
\small
\centering
\renewcommand\arraystretch{1.1}
\caption{Ablation experiments on different clustering methods in PG-CIDL. The baseline model is vanilla Swin Transformer.\label{tab:table2}}
\begin{tabular}{lcc}
\Xhline{0.8pt}
Models (AMU-CSCC)& ACC& AUC\\ \hline
MFC-MIL&0.8962&\underline{0.9723}\\
ACMIL-MHA &\underline{0.9057}&0.9701\\
 baseline & \underline{0.8868} & 0.9778 $\uparrow$ \\ \hline
PG-CIDL w/o PSD-LFG& 0.8962 $\uparrow$
& 0.9641 $\downarrow$\\
PG-CIDL w/o CID & 0.8868& 0.9825 \\
PG-CIDL & \textbf{0.9434} $\uparrow$& \textbf{0.9859} $\uparrow$\\
\Xhline{0.8pt}
\end{tabular}
\end{table}

\begin{table}[!t]
\small
\centering
\renewcommand\arraystretch{1.1}
\caption{Ablation experiments on different metrics in CID.\label{tab:table3}}
\begin{tabular}{lcccc}
\Xhline{0.8pt}
\multirow{2}*{Metrics in CID}& \multicolumn{2}{c}{AMU-CSCC} & \multicolumn{2}{c}{CAMELYON16} \\ \cmidrule(r){2-3} \cmidrule(r){4-5} 
~ & ACC & AUC & ACC & AUC\\ \hline
Cosine distance & 0.9151 & 0.9776 & 0.6857 & 0.7771\\
Hellinger distance & 0.9245 & 0.9795 & \underline{0.8857} & \underline{0.9151} \\
JS divergence & \underline{0.9434} & \underline{0.9849} & 0.8571 & 0.8882 \\ \hline
\textbf{KL divergence (ours)} & \textbf{0.9434} & \textbf{0.9859} & \textbf{0.8857} & \textbf{0.9282}\\
\Xhline{0.8pt}
\end{tabular}
\end{table}

\subsection{Ablation Experiments}
To validate the effectiveness of components proposed in PG-CIDL, we conducted extensive ablation experiments. In Table \ref{tab:table2}, we choose the baseline model as a fully supervised mean-pooling framework, which outperforms the second-best MFC-MIL on AMU-CSCC with 0.55\% AUC improvement but ACC drops 1.89\% compared to ACMIL-MHA. In general, It suggests that end-to-end learning can learn task-related feature representation that leads to better classification performance.

\begin{figure*}[!t]
\centering
{\includegraphics[width=7in]{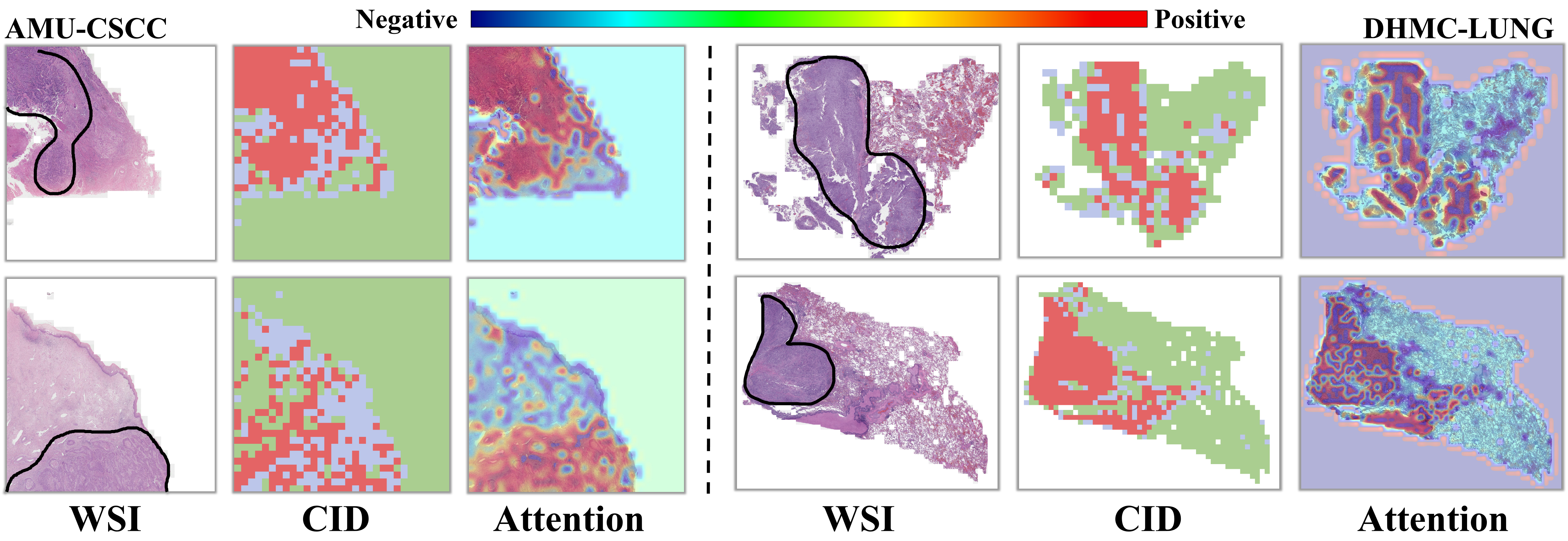}}
\hfil
\caption{Visualization results on AMU-CSCC and DHMC-LUNG. The first column is original WSI with pathologist-annotated tumor boundaries. The second column is the labeled regions after CID, where red, blue, green denote factor tumor, environment and background respectively. The last column is the attention heatmap corresponding to its category.}
\label{fig:fig5}
\end{figure*}

\begin{figure*}[!t]
\centering
{\includegraphics[width=7in]{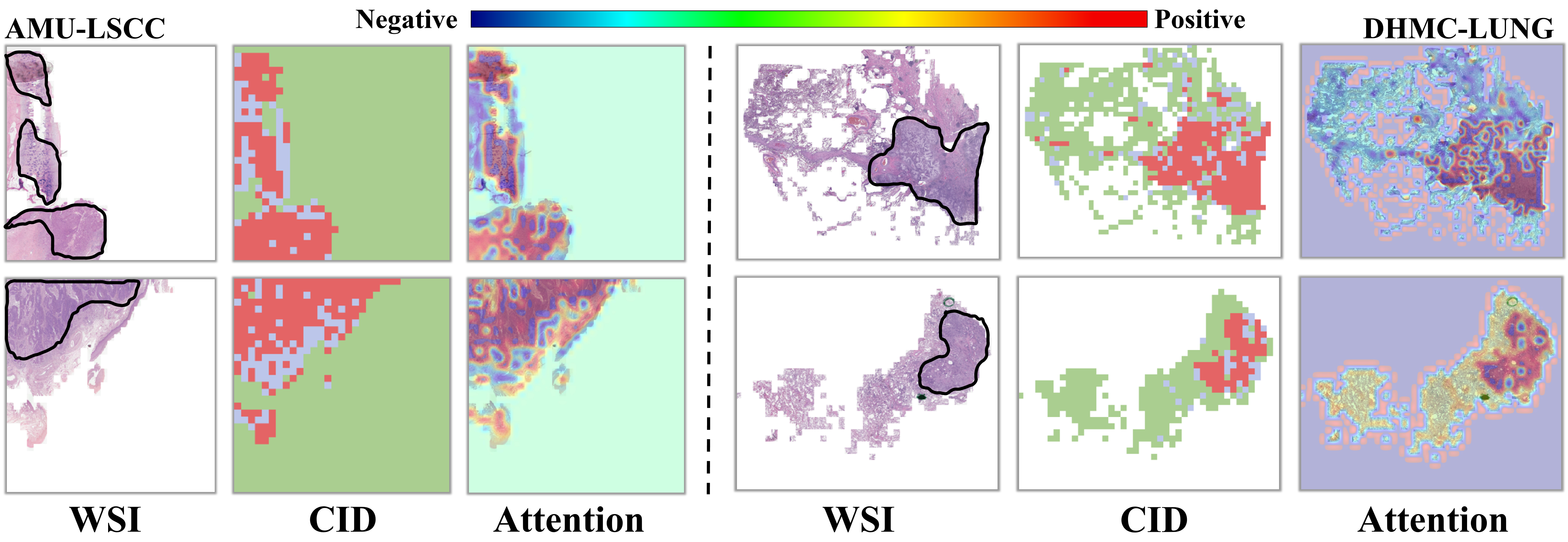}}
\hfil
\caption{Extra visualization results on AMU-LSCC and DHMC-LUNG.}
\label{fig:vis_lscc}
\end{figure*}

\begin{figure*}[!t]
\centering
{\includegraphics[width=7.3in]{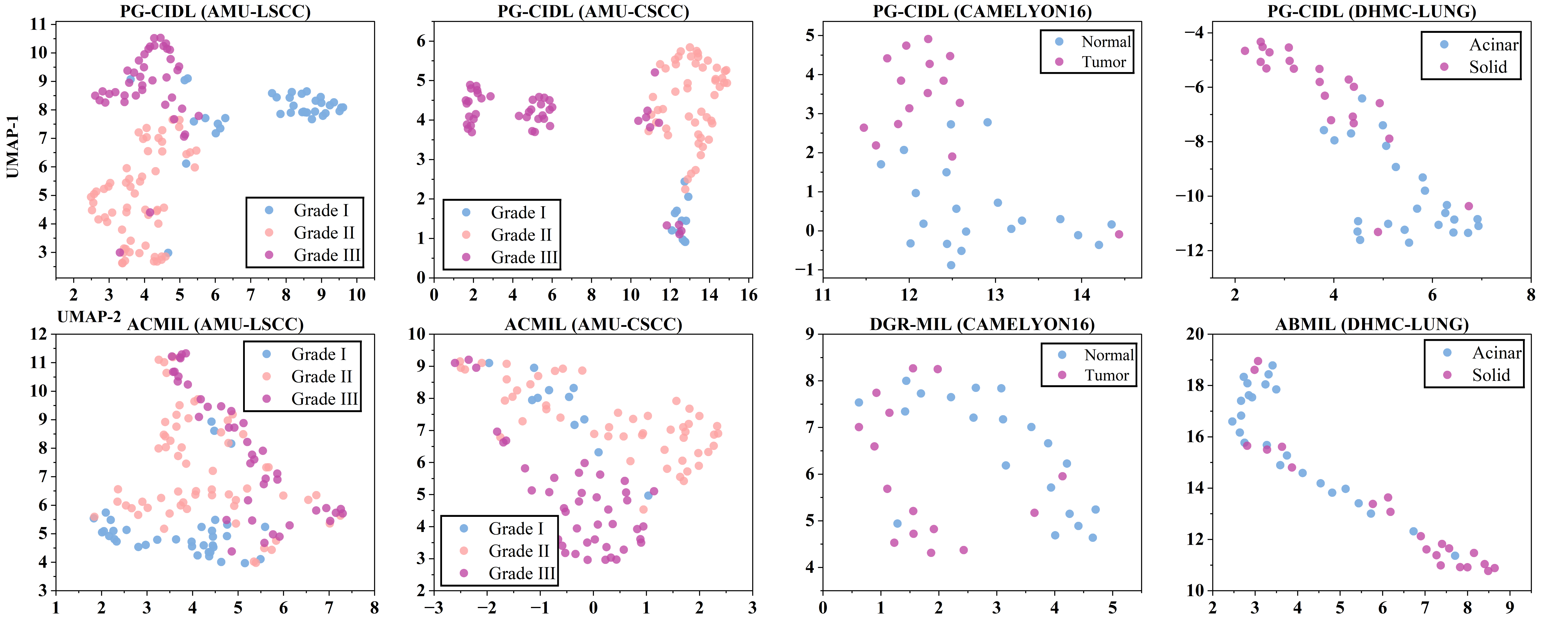}}
\hfil
\caption{2-D feature representation with spatial disentanglement on multicentre datasets.}
\label{fig:2d_cluster}
\end{figure*}

We simply consider three factors due to the motivation that for most WSI diagnosis tasks, the main variability is between tumor, its microenvironment and background. Larger numbers of groups are needed for more complex tasks \cite{song2024fasial}. We validate PSD-LFG by replacing with L2-KMeans, we observe a 0.94\% improvement in ACC and a 1.37\% drop in AUC. Since instance grouping is the basis for counterfactual inference, an inaccurate and entangled grouping result may introduce epistemic errors into the causal graph and disentangling process. Incorporating PSD-LFG, PG-CIDL then ensures a robust reasoning process with disentangled and refined representation, leading to optimal classification performance. Besides, we test the performance of using grouping strategy alone with fixed attention weighting. It shows that our PG-CIDL degrades without causal effects re-weighting from CID. Instead, combining these two components improves both ACC and AUC.

Moreover, Table \ref{tab:table3} reports the performance of different metrics employed in CID module. It indicates that cosine distance is the least suitable metric. As a symmetric version of KL, the JS divergence achieves comparable performance with only a 0.1\% drop in ACC on the AMU-CSCC. Overall, these results demonstrate that incorporating KL divergence in CID module yields superior performance.

To further validate the generalization capability of PG-CIDL, we performed end-to-end training on selected 7 SOTA MIL models for fairer comparison. In Table \ref{tab:e2e}, these models achieve slightly better performance compared to two-stage training. However, our proposed PG-CIDL with disentangled representation still outperform these MIL models. It especially improved the second-best mACC by 8.69\%  and AUC by 4.46\%.

\begin{table}[t]
\small
\centering
\renewcommand\arraystretch{1.2}
\caption{End-to-end performance comparison of PG-CIDL and other SOTA models on LSCC.\label{tab:e2e}}
\begin{tabular}{cccc}
\hline
\Xhline{0.8pt}
End-to-End & mAcc & WF1 & AUC     \\ \hline
ABMIL(Linear) & 0.7826 & 0.7791 & 0.9120\\
TransMIL  & 0.8188 & 0.8205 & 0.9116\\
DGR-MIL  &0.8188  & 0.8165 & 0.9273 \\
ACMIL-MHA  & 0.8406 & 0.8414 & 0.9252\\
ILRA-MIL  & 0.8116 & 0.8119 & 0.9198 \\
RRT-MIL  & 0.7826 & 0.7815 & 0.9038 \\
MFC-MIL& 0.8116& 0.8112 & 0.9115  \\
\textbf{PG-CIDL (ours) }& \textbf{0.9275} & \textbf{0.9264} & \textbf{0.9719}  \\
\Xhline{0.8pt}
\hline
\end{tabular}
\end{table}

\subsection{Visualization of Model Interpretability}
To validate whether our model managed to identity instance features into three disentangled factors, we provide visualization results of our PG-CIDL on AMU-CSCC, AMU-LSCC and DMHC-LUNG. In Figure \ref{fig:fig5} and Figure \ref{fig:vis_lscc}. We can see that both disentangled factors and attention heatmaps highlight the area that are highly consistent with pathologists' annotations. More specifically, the clustering and labeling results from CID reflect the model’s active interpretability, as it disentangles semantically ambiguous features and explicitly groups regions with tumor, microenvironment and background before final prediction, mirroring a real-world diagnostic workflow. Meanwhile, the attention heatmaps highlight the most contributive regions for classification, revealing the model’s passive interpretability.

Overall, these two kinds of interpretability demonstrate that PG-CIDL not only learns more interpretable and task-relevant representations, but also aligns well with clinical reasoning, offering a comprehensive explanation of predictions. It also implies a potential paradigm for weakly supervised semantic segmentation.

\begin{figure}[!t]
\centering
{\includegraphics[width=3.5in]{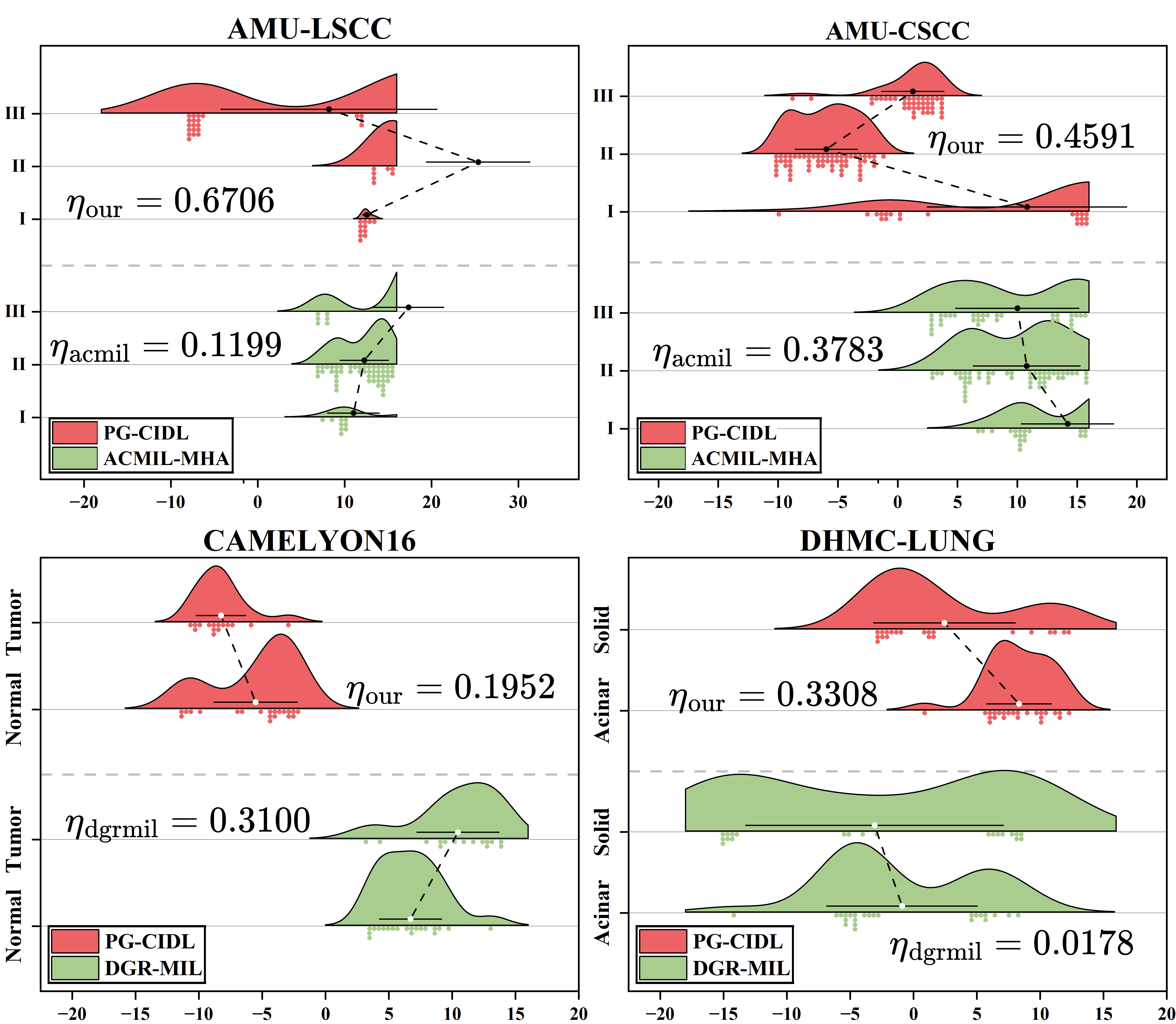}}
\hfil
\caption{ANOVA plots on four datasets where $\eta_*$ denote the value of effect size.}
\label{fig:fig6}
\end{figure}

\subsection{Capability of Feature Representation}
We utilize UMAP \cite{becht2019umap} to visualize feature representation in 2-D space. In Figure \ref{fig:2d_cluster}, we can see that our proposed PG-CIDL show less entanglement within feature distributions, while other sub-optimal models without spatial disentanglement show more feature overlaps. It not only implies that PG-CIDL disentangles feature spatially, but it also has superior ability to distinguish feature related to corresponding grading results.

An effective classification model should be able to capture and fit the differences among pathological patterns. To validate this assumption, we visualize the final layer feature representation of each model using UMAP in 1-D space. Figure \ref{fig:fig6} shows the analysis of variance (ANOVA) with effect size metrics ($\eta^2$).

\begin{figure}[!t]
\centering
{\includegraphics[width=3.8in]{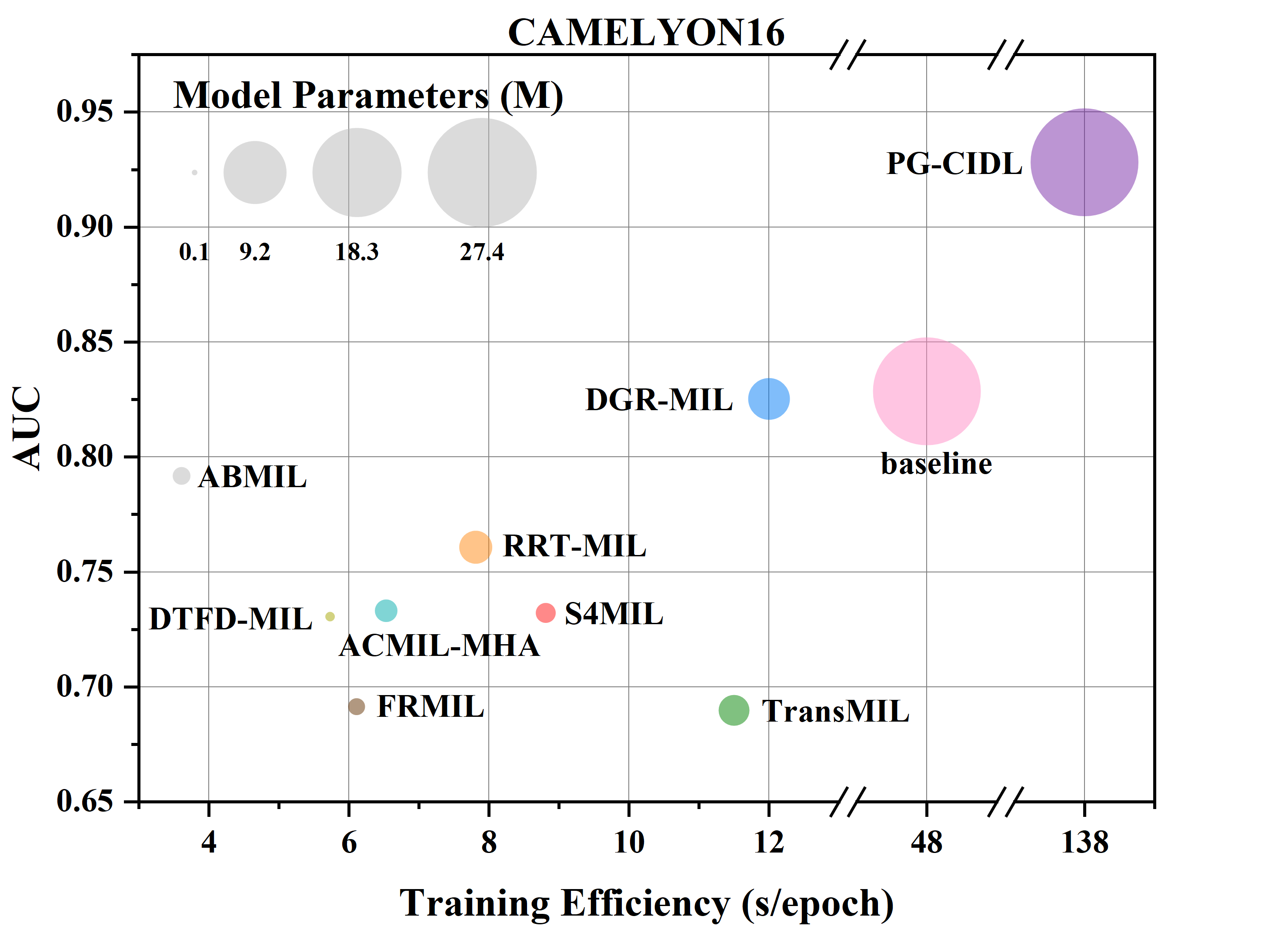}}
\hfil
\caption{Model complexity vs. performance.}
\label{fig:fig7}
\end{figure}

The effect size in ANOVA quantifies the proportion of variance in the dependent variable explained by the group labels. A larger effect size suggests a stronger ability of the model to capture and distinguish differences between classes. In Figure \ref{fig:fig6}, PG-CIDL models the distinct distributions within different categories, while ACMIL and DGR-MIL tend to overlap. It has larger effect size 0.6706, 0.4591, 0.3308 compared to the second-best models on AMU-CSCC, AMU-LSCC and DHMC-LUNG, respectively. Although DGR-MIL on CAMELYON16 exhibit effect size 1.5 times larger, PG-CIDL still remains a considerable amount of $\eta^2$. Therefore, our proposed PG-CIDL exhibits better feature feature representation ability in general, which directly implies that our model is capable of address all three aspects of entanglement to obtain more explainable and generalized representation.

\subsection{Model Complexity Analysis}
End-to-end learning for WSI analysis is considered to be computationally intensive, accordingly receiving limited exploration. However, our findings suggest that its actual computational demand has been overestimated. In Supplementary Material, we present detailed computational specs. In Figure \ref{fig:fig7} we present comparison of complexity vs. performance on CAMELYON16 of batch size 1.

Specifically, our PG-CIDL contains 27.50M parameters significantly more than existing models, which range from 0.199M (DTFD-MIL) to 4.075M (DGR-MIL). This leads to a longer training time per epoch for PG-CIDL (138s), compared to 12.0s for DGR-MIL, the most computationally demanding among the SOTA models. Although our PG-CIDL approach increases training time ten times larger, the computational cost is relatively accepted. It allows the model to learn highly task-specific feature representations, resulting in better classification accuracy and enhanced interpretability.

\section{Conclusion}
WSI-based pathological grading is the golden standard for clinical diagnosis and prognosis. As one of the main-stream frameworks for WSI classification, multi-instance learning models suffer from weak interpretability and generalizability caused by spatially, semantically and decision entangled representations. Guided by a prior structural causal model, we proposed a three-phase disentangled representation learning framework PG-CIDL, followed by positive semi-definite latent factor grouping, counterfactual inference-based cluster-reasoning instance disentangling and instance effect re-weighting. In this case, our PG-CIDL decomposes the entangled features into three causally relevant factors, namely tumor, microenvironment, and background, and refines them to derive a more interpretable WSI representation. Extensive experiments on multicentre datasets demonstrate that our PG-CIDL not only achieves superior classification performance, but exhibits strong capability of disentangled feature representation, interpretability and generalizability. Our study also explored end-to-end learning for WSIs. Unlike conventional two-step models that rely on offline features or separate clustering pipelines, our unified framework optimizes feature across all stages. This leads to more task-relevant and interpretable representations.

\bibliographystyle{IEEEtran}
\bibliography{IEEEabrv,my_ref}

\end{document}